%% file: main.tex
\documentclass[lettersize,journal]{IEEEtran}
\usepackage{amsmath,amsfonts}
\usepackage{algorithmic}
\usepackage{array}
\usepackage{textcomp}
\usepackage{stfloats}
\usepackage{url}
\usepackage{verbatim}
\usepackage{graphicx}
\usepackage{booktabs,tabularx,array,multirow}
\usepackage{subcaption}
\usepackage{hyperref}
\usepackage{cleveref}
\usepackage[table]{xcolor}

\def\BibTeX{{\rm B\kern-.05em{\sc i\kern-.025em b}\kern-.08em
    T\kern-.1667em\lower.7ex\hbox{E}\kern-.125emX}}
\usepackage{balance}
\usepackage{xcolor}

\usepackage{float}
\usepackage{caption}
\crefname{figure}{Fig.}{Figs.}
\Crefname{figure}{Fig.}{Figs.}

\begin{document}
\title{\textbf{RoboIRS}: Inference-Time \textbf{I}nternal \textbf{R}epresentation \textbf{S}teering for Generalist Robot Policies}
\author{
Jiuzhou Lei$^{1}$,
Chang Liu$^{1}$,
Dayou Li$^{1}$,
Zhiyuan Zhang$^{2}$,
Xiao Liang$^{3}$,
Yu She$^{2}$,
Zhiwen Fan$^{4}$,
Minghui Zheng$^{1}$
\thanks{This work was partially supported by the U.S. National Science Foundation
under Grant Nos. 2422826, 2423068, and 2527316.}
\thanks{$^{1}$ Jiuzhou Lei, Chang Liu, Dayou Li, and Minghui Zheng are with the
J. Mike Walker '66 Department of Mechanical Engineering, Texas A\&M University,
College Station, TX 77843, USA.
{\tt\small Emails: \{jiuzl, changliu.chris, dayou519, mhzheng\}@tamu.edu}}
\thanks{$^{2}$ Yu She and Zhiyuan Zhang are with the Edwardson School of
Industrial Engineering, Purdue University, West Lafayette, IN, USA.
{\tt\small Emails: \{yushe, zhan5570\}@purdue.edu}}
\thanks{$^{3}$ Xiao Liang is with the Zachry Department of Civil and
Environmental Engineering, Texas A\&M University, College Station, TX 77843, USA.
{\tt\small Email: xliang@tamu.edu}}
\thanks{$^{4}$ Zhiwen Fan is with the Department of Electrical and Computer
Engineering, Texas A\&M University, College Station, TX 77843, USA.
{\tt\small Email: zhiwenfan@tamu.edu}}
}

\markboth{}%
{}

\maketitle

\begin{abstract}
Vision-language-action (VLA) and world-action models (WAMs) often degrade under out-of-distribution task variations despite retaining partial task capability.
To recover such capability, we propose RoboIRS, an inference-time internal representation steering method that uses successful and failed rollouts to train linear classifiers, select outcome-relevant intervention locations, and derive task-specific steering directions without updating policy parameters.
On 15 simulation tasks with a frozen $\pi_{0.5}$ policy, RoboIRS improves the average success rate from 44.4\% to 66.2\%, outperforming alternative inference-time intervention baselines while adding little inference time.
We further validate RoboIRS on real-robot manipulation using the same $\pi_{0.5}$ policy and demonstrate its applicability to a world-action model Cosmos Policy, where the average success rate improves from 35.4\% to 55.4\%.
These results show that directly steering internal robot-policy representations can improve the performance of robot policies at inference time. Project website is available at \href{https://rollingoat.github.io/roboirs/} {\textcolor{blue}{https://rollingoat.github.io/roboirs/}}.
\end{abstract}


\input{tex/introduction}

\input{tex/related-work}

\input{tex/proposed-approach.tex}

\input{tex/results-and-analysis.tex}
\input{tex/conclusion.tex}
\section*{Acknowledgments}
We also thank Hao Wang and Nuo Chen for their suggestions during discussion.
As listed in submission requirements, we disclose that AI tools like codex and claude are used to generate code for experiments and implementation which are reviewed by authors. 
\bibliographystyle{IEEEtran}
\bibliography{ref}

\end{document}

%% file: tex/introduction.tex
\section{Introduction}
\label{sec:introduction}

Vision-language-action (VLA) models and world-action models (WAMs) have recently emerged as promising architectures for robotic manipulation by integrating visual perception, language-conditioned task understanding, and action generation with future prediction in a unified policy. However, their performance often degrades on out-of-distribution (OOD) inputs, such as unseen scenes, novel object layouts, and new object appearances \cite{zhou2025libero, fei2026libero}. Importantly, this degradation does not necessarily imply complete incapability: a policy may still succeed on a subset of these OOD tasks, albeit at a lower success rate. This suggests that the policy may retain some of the internal computations required for successful execution even when its behavior becomes unreliable.

\begin{figure}[!t]
      \centering
      \includegraphics[width=0.8\columnwidth]{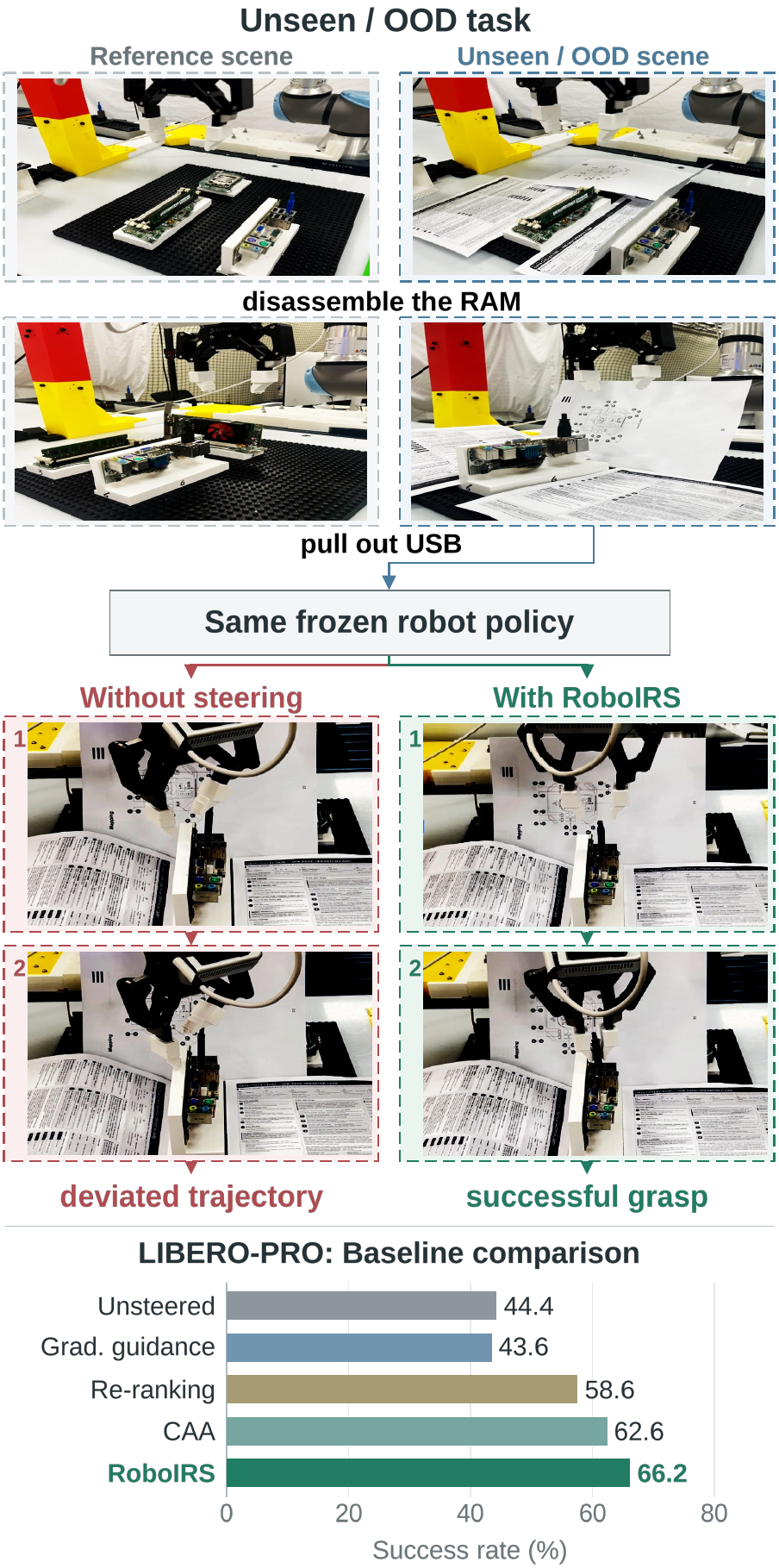}
      \caption{Real-robot setup under visual distribution shift (top) 
      and baseline comparison (bottom).}
      \label{fig:teaser}
  \end{figure}

A similar phenomenon has already been observed in LLMs, which may produce both truthful and untruthful responses despite being queried under similar conditions.
This inconsistency suggests that the model may already contain internal representations associated with the desired behavior, even when they are not reliably expressed in its outputs \cite{li2023inference}.
Recent progress in mechanistic interpretability and activation steering for large language models provides evidence that internal representations can be causally manipulated to alter model behavior.
Here, internal representations refer to intermediate output that encode information during model computation; in transformer-based models, these may include residual-stream activations passed between layers, as well as activations associated with individual neurons, attention heads, or other localized model components \cite{zhang2026locate}.
Prior work has shown that high-level behaviors such as truthfulness, refusal, sentiment, and task execution can be localized in hidden representations and modified through activation-space interventions. Li et al.~\cite{li2023inference} steer LLMs toward truthful answers by intervening on truth-related activations, while Turner et al.~\cite{turner2023steering} and Rimsky et al.~\cite{rimsky2024steering} show that contrastive activation directions can control generation behavior without retraining the model. Zou et al.~\cite{zou2023representation} frame such methods as representation engineering, where internal activations are treated as controllable objects for modifying model behavior.

Motivated by the findings in LLM, can we similarly steer the internal representations of a robot policy at inference time to improve its performance on OOD tasks?
Recent work has already extended mechanistic interpretability to vision-language-action (VLA) models. Grant et al.~\cite{grant2026not} and Swann et al.~\cite{swann2026sparse} use sparse autoencoders to identify interpretable residual-stream features associated with robot behaviors, but do not demonstrate how these features can improve downstream policy performance. Häon et al.~\cite{haon2025mechanistic} and Buurmeijer et al.~\cite{buurmeijer2026observing} go further by intervening on internal representations to steer behaviors such as motion speed and end-effector height; however, their evaluations are limited to a small set of predefined behavioral attributes rather than general task performance. Hong et al.~\cite{hong2026steering} instead use an LQR controller to steer the diffusion transformer’s hidden activations such that the resulting activations lead to robust robot behaviors and require constructed contrastive residual-stream pairs from matched nominal and perturbed observations to provide target activation levels.

In this work, we aim to improve the performance of a frozen robot policy under distribution shift through inference-time steering of its internal representations.
Rather than retraining the policy, we use these outcome-labeled activations to identify representations associated with successful and failed rollouts and intervene on the policy at inference time using a steering vector.
The goal is to bias an already capable but unreliable policy toward internal states that are more predictive of successful action selection. 
Inspired by \cite{li2023inference}, we first train a linear classifier to classify the internal representations and then use the classifier weight as steering vectors to change the internal representations such that the predicted actions are more favorable of successful rollouts.
We evaluate the proposed approach on two representative robot policies: the vision-language-action model \(\pi_{0.5}\)~\cite{intelligence2025pi} and the world-action model Cosmos Policy~\cite{kim2026cosmos}. 
We conduct experiments in both simulation and real-world robotic settings to assess whether inference-time internal representation steering improves task success rates under distribution shift.

The main contributions of this work are as follows.
\begin{itemize}
    \item We propose RoboIRS, which adapts the internal representation-steering paradigm from LLMs to robot policies and enables inference-time intervention on frozen policies using steering vectors derived from the internal representations of successful and failed rollouts.
    \item We conduct simulation and real-robot experiments showing that RoboIRS improves the average SR of a fully trained $\pi_{0.5}$ policy under distribution shift from $44.4\%$ to $66.2\%$, outperforms alternative inference-time intervention baselines, improves real-robot manipulation performance, and is also effective on Cosmos Policy despite its different training recipe and model architecture, increasing the average SR from $35.4\%$ to $55.4\%$.
\end{itemize}

%% file: tex/related-work.tex
\section{Related Work}
\label{sec: related work}

\subsection{Mechanistic Interpretability and Steering in LLMs}

Mechanistic interpretability increasingly studies how high-level concepts are represented in neural activations and whether these representations can be causally manipulated~\cite{zhang2026locate,geiger2025causal}.
Many intervention methods build on the observation that properties such as truthfulness and sentiment can be represented by linear directions or low-dimensional subspaces~\cite{lieanrRepresentationHypothesis,marks2023geometry,tigges2023linear}.
Activation-steering methods exploit this structure by constructing directions from contrastive examples or demonstrations and injecting them into hidden states to control behavior without retraining~\cite{turner2023steering, rimsky2024steering, hendel2023context, todd2023function}.
Such interventions have been applied to truthfulness, hallucination, and refusal behavior~\cite{li2023inference,zhang2024truthx,wang2025adaptive,arditi2024refusal}.

\subsection{Mechanistic Interpretability and Internal Representation Steering of Robot Policy Models}

Recent robotics work has begun to apply mechanistic interpretability techniques to characterize the internal representations of VLA policies, with several studies further showing that the discovered features can serve as causal handles for behavior steering.
Khan et al.~\cite{khan2025controlling} use sparse autoencoders to identify action-related latent directions from contrastive action pairs and show that these directions can be manipulated to steer VLA action outputs without retraining.
Grant et al.~\cite{grant2026not} conduct a cross-architecture mechanistic study using activation injection, sparse autoencoders, and linear probes to analyze how visual, language, and action-generation pathways contribute to VLA behavior.
Swann et al.~\cite{swann2026sparse} use sparse autoencoders to identify interpretable features in VLA hidden states, distinguishing transferable motion primitives from episode-specific memorization, and causally validate selected features through steering experiments.

Several VLA papers move from feature discovery to explicit control of robot behaviors.
Buurmeijer et al.~\cite{buurmeijer2026observing} formalize feature observability and controllability in VLA models and use lightweight linear interventions to steer encoded features during closed-loop execution.
Häon et al.~\cite{haon2025mechanistic} identify sparse semantic directions in transformer activations and use those directions to control policy behavior without retraining.
Mitra et al.~\cite{mitra2025mechanistic} identify task-relevant attention heads and selectively update them, framing policy adaptation as targeted modification of relevant internal mechanisms.
COAST~\cite{miao2026contrastive} constructs success- and failure-associated conceptor from rollout hidden states to steer VLA activations toward success-critical representations at inference time. 
COAST is evaluated on imperfect policies, such as early fine-tuning checkpoints, using tasks drawn from the original benchmark distribution, whereas RoboIRS focuses on recovering the performance of fully trained frozen policies under unseen task perturbations.
Hong et al.~\cite{hong2026steering} take a control-based approach to representation steering, constructing contrastive directions from paired nominal and perturbed inputs and using them to define robustness-relevant latent feature subspaces and target feature setpoints.
RoboIRS instead derives its intervention directly from successful and failed rollouts without requiring matched nominal and perturbed observation pairs.

\subsection{Inference-Time Intervention of Robot Policies}

Another line of work improves robot policies at inference time through action selection or generative guidance. 
V-GPS~\cite{nakamoto2025steering} samples actions from a frozen policy and re-ranks them with an offline-RL value function, while RoboMonkey~\cite{kwok2025robomonkey}, MG-Select~\cite{jang2026verifierfree}, and RoVer~\cite{dai2025rover} select or refine VLA action candidates using learned verifiers, intrinsic confidence, or process rewards. 
Related methods intervene directly during action generation: value-guided diffusion~\cite{zhang2024valueguided} biases diffusion trajectories with value gradients. Guided Action Flow~\cite{yang2026guided} keeps the VLA frozen and uses gradients from a critic trained on successful and failed rollouts to steer flow-matching action generation toward higher-return actions. 
PPGuide~\cite{wang2026ppguide} trains a performance predictor and uses its guidance signal to steer pretrained diffusion policies away from predicted failure modes during inference.

%% file: tex/proposed-approach.tex
\section{Proposed Approach}
\label{sec:proposed-approach}



\subsection{Model Architecture and Internal Representation Definitions}


\begin{figure*}[!t]
      \centering
      \includegraphics[width=0.9\textwidth]{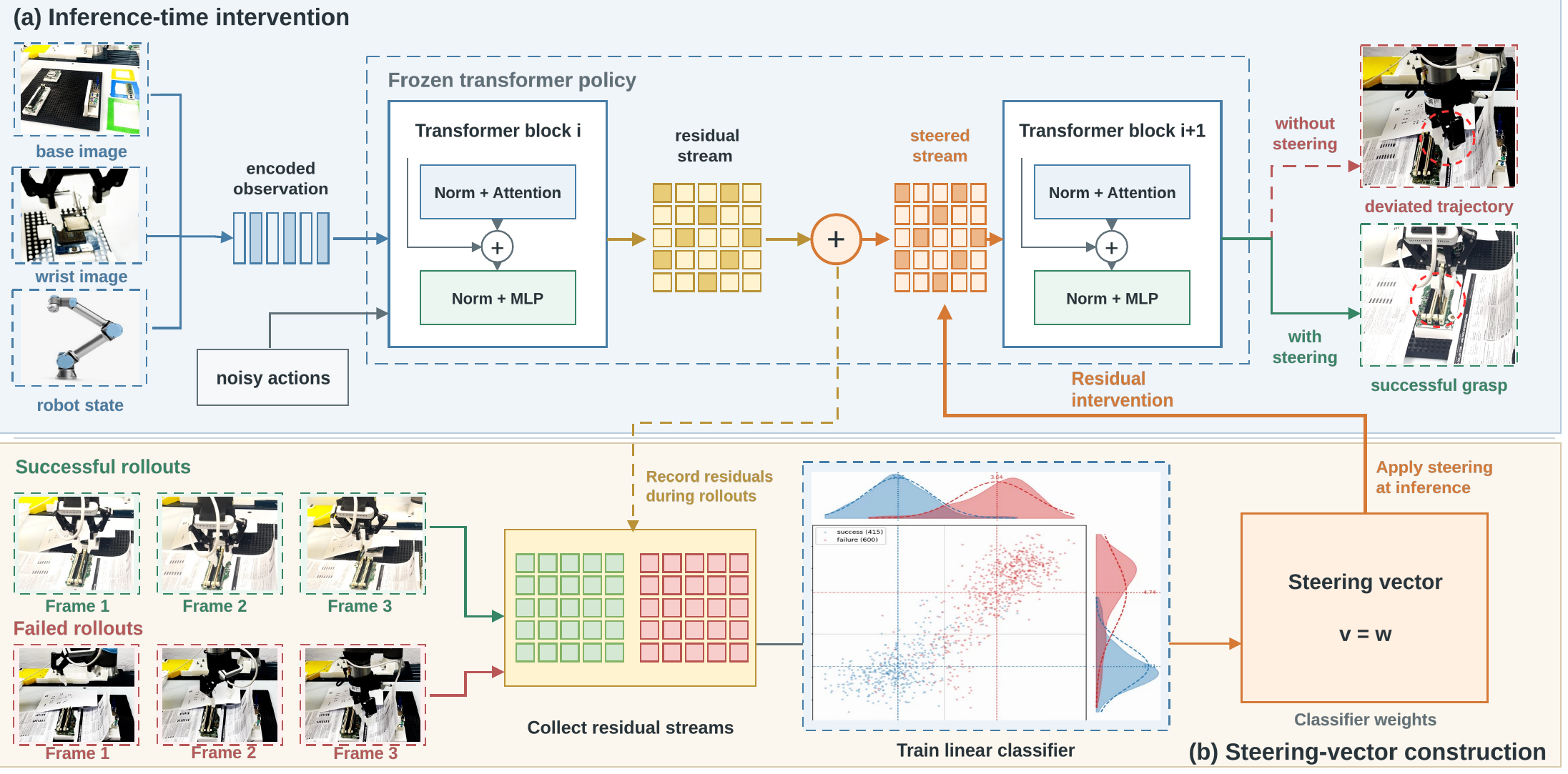}
      \caption{
      Overview of the proposed RoboIRS framework.
      (a) RoboIRS intervenes between transformer blocks by adding a scaled steering vector to the selected residual-stream representation without updating the policy parameters.
      (b) To construct the steering vector, residual-stream representations are collected from successful and failed rollouts and used to train a linear classifier. The classifier weight provides the steering direction that is injected into the residual stream during inference.
      }
      \label{fig:proposed_approach}
  \end{figure*}

To explain how steering is applied to internal representations, we first briefly describe the model architecture and define the notation used throughout this section.
Since $\pi_{0.5}$ is the primary VLA model used in our experiments, we use it as a running example, although the formulation should apply more generally to transformer-based robot policies.

The $\pi_{0.5}$ VLA consists of a pretrained PaliGemma vision-language backbone and a smaller transformer-based action expert.
The action expert, whose internal representations we steer, contains $L=18$ transformer layers.
Actions are generated using flow matching.
At each environment timestep $t$, the policy starts from a noisy action chunk and iteratively refines it over $N$ denoising steps.
Let $\mathbf{X}_{t,n} \in \mathbb{R}^{T_a \times d_a}$ denote the noisy action chunk at denoising step $n$, where $T_a$ is the action-chunk length and $d_a$ is the dimensionality of one robot action.
Each row $\mathbf{x}_{t,n,j} \in \mathbb{R}^{d_a}$ therefore corresponds to the $j$-th action in the chunk.

The action expert projects the action chunk into its hidden dimension $d$, producing $T_a$ action tokens represented by $\mathbf{R}_{0,t,n} \in \mathbb{R}^{T_a \times d}$.
For $\pi_{0.5}$, $d=1024$.
These tokens are processed through the $L$ transformer blocks while conditioned on the VLM context, after which the action expert predicts the flow velocity used to update the action chunk.

Let $\mathbf{R}_{\ell,t,n} \in \mathbb{R}^{T_a \times d}$ denote the residual stream entering transformer layer $\ell$.
Omitting $t$ and $n$ for clarity, each transformer block updates the residual stream as

\begin{equation}
    \mathbf{R}'_{\ell}
    =
    \mathbf{R}_{\ell}
    +
    \operatorname{Attention}\!\left(
        \operatorname{Norm}(\mathbf{R}_{\ell})
    \right),
    \label{eq:attention}
\end{equation}
followed by

\begin{equation}
    \mathbf{R}_{\ell+1}
    =
    \mathbf{R}'_{\ell}
    +
    \operatorname{MLP}\!\left(
        \operatorname{Norm}(\mathbf{R}'_{\ell})
    \right).
    \label{eq:mlp}
\end{equation}
The representation of action token $j$ is denoted by $\mathbf{r}_{\ell,t,n,j} \in \mathbb{R}^{d}$, corresponding to the $j$-th row of $\mathbf{R}_{\ell,t,n}$.
These token-level residual-stream representations are the intervention sites used for steering.

\subsection{Steering Vector}

\textbf{Data Collection.}
Constructing the steering vectors requires residual streams from both successful and failed rollouts.
For rollout $i$ with $T_i$ environment timesteps, we record $\mathbf{R}_{\ell,t,n}^{(i)}$ for every timestep $t$, transformer layer $\ell$, and denoising step $n$.
The resulting activation tensor is $\mathcal{R}^{(i)} \in \mathbb{R}^{T_i \times L \times N \times T_a \times d}$.
For example, a 30-step rollout with $\pi_{0.5}$ produces a tensor of shape $(30,18,10,10,1024)$, corresponding to 30 environment timesteps, 18 transformer layers, 10 denoising steps, 10 action tokens, and a hidden dimension of 1024.

We collect up to 10 successful and 10 failed rollouts per task to reduce underdetermination when fitting the linear classifier while keeping data-collection cost manageable.
For tasks with very high or very low success rates, fewer examples of the minority outcome may be collected to avoid an excessive number of rollouts.

\textbf{Contrastive Difference as Steering Vector.}
Inspired by representation-steering methods for LLMs \cite{rimsky2024steering, li2023inference}, we consider two approaches for constructing steering vectors.
The first uses the mean contrastive difference between residual representations from successful and failed rollouts.
For each layer $\ell$, denoising step $n$, and action-token position $j$, we compute

\begin{equation}
\mathbf{v}_{\ell,n,j}
=
\frac{1}{|\mathcal{D}_{s}|}
\sum_{(i,t)\in\mathcal{D}_{s}}
\mathbf{r}^{(i)}_{\ell,t,n,j}
-
\frac{1}{|\mathcal{D}_{f}|}
\sum_{(i,t)\in\mathcal{D}_{f}}
\mathbf{r}^{(i)}_{\ell,t,n,j},
\label{eq:contrastive_vector}
\end{equation}
where $\mathcal{D}_{s}$ and $\mathcal{D}_{f}$ denote the sets of timestep samples from successful and failed rollouts, respectively.
Each steering vector $\mathbf{v}_{\ell,n,j} \in \mathbb{R}^{d}$ therefore corresponds to a specific transformer layer, denoising step, and action-token position.

\textbf{Linear Classifier Weight as Steering Vector.}
The second approach uses the weight of a linear classifier as the steering direction, following analogous approaches in LLMs \cite{li2023inference}.
For each task, transformer layer, and action-token position, we train a linear classifier to distinguish success-like from failure-like residual streams.
The residual stream's label from each timestep is binary: all timesteps of successful rollouts are 0, and a failed rollout's timesteps become 1 from the first step where its end-effector leaves the space swept by the successful trajectories.

The failure classifier is defined as
\begin{equation}
P(y=1 \mid \mathbf{x})
=
\sigma\!\left(\mathbf{w}^{\top}\mathbf{x}+b\right),
\label{eq:failure_classifier}
\end{equation}
where $y=1$ denotes failure, $\mathbf{w}\in\mathbb{R}^{d}$ is the classifier weight, and $b\in\mathbb{R}$ is the bias.
We use $\mathbf{w}$ as the corresponding steering direction.
Thus, each layer--token pair has its own classifier and steering vector.
Although this requires training multiple classifiers per task, both the classifiers and datasets are small; in our experiments, all classifiers can be trained in under three minutes using multiple CPUs.

As noted in \cite{grant2026not}, linear probes trained on residual streams from the action expert achieve high area under the ROC curve (AUC) for predicting task success.
Consistent with this finding, our classifiers also achieve good AUC when distinguishing representations from successful and failed rollouts. 
The average validation AUC is 0.88 for all trained classifiers.

\subsection{Intervention}

One limitation of using the mean contrastive difference alone is that it does not indicate which action-token position or transformer layer should be steered.
One could perform a sweep over intervention sites and intervention strength, as in \cite{rimsky2024steering}, to identify the most effective place for intervention.
Such sweeps are relatively inexpensive for LLMs, where performance can often be evaluated with a single question--answer interaction.
In robotics, however, each candidate typically requires multi-step policy inferences, making an exhaustive search over intervention sites prohibitively expensive.
Instead, we use the linear classifier's AUC on held-out validation rollouts to select the steering location.
From the collected successful and failed rollouts, we reserve a subset from each outcome as validation data and use the remaining rollouts for classifier training.
For each layer $\ell$, we select the action-token position with the highest failure-detection AUC to steer. This choice is further explained in \Cref{sec:q2_baseline_comparison}.

Once the steering location is determined, the selected residual stream at an action token is modified as

\begin{equation}
\mathbf{r}_{\mathrm{steered}}
=
\mathbf{r}
+
s\alpha\frac{\|\mathbf{r}\|}{\|\boldsymbol{v}\|}\mathbf{v},
\qquad
s\in\{-1,+1\},
\label{eq:steering_intervention}
\end{equation}
where $\alpha \geq 0$ controls the steering strength and $s$ determines the steering direction. 
Since the residual-stream magnitude varies across layers, we scale the intervention by $\frac{||\mathbf{r}||}{||\mathbf{v}||}$ so that the steering strength is proportional to the magnitude of the residual stream at each layer.

The sign of the intervention is selected empirically rather than assuming that the steering vector is to pull residual streams from failure-relevant space to success-relevant space.
The linear representation hypothesis \cite{lieanrRepresentationHypothesis} suggests that high-level concepts can be represented as directions in a model's representation space. 
From this view, we interpret $\mathbf{v}$ as a task-relevant latent direction rather than a  direction pointing from failure relevant latent space to success relevant spaces.
We hypothesize that successful behavior may require an appropriate level of the latent factor encoded by $\mathbf{v}$. 
For a task in which this factor is under-activated, steering along $+\mathbf{v}$ may improve performance, whereas a task in which it is over-activated may benefit from $-\mathbf{v}$.
We therefore first evaluate one steering sign and test the opposite sign only when the sign shows a monotonic performance decrease as shown in \Cref{fig:steering_strength}, avoiding twice the rollout cost for every task.
Importantly, a linearly discriminative direction need not be a causal intervention direction \cite{lieanrRepresentationHypothesis}. 
Therefore, particularly for the classifier-derived vectors, the direction may encode information that is predictive but not causally useful to change robot behaviors.

%% file: tex/results-and-analysis.tex
\section{Results and Analysis}
\label{sec:results_analysis}

We structure our experiments around the following questions:

\begin{itemize}
    \item \textbf{Q1:} Does steering the internal representations improve task success rates in both simulation and on the real-robot experiment?
    \item \textbf{Q2:} Does the proposed approach outperform alternative inference time intervention methods?
    \item \textbf{Q3:} Can the proposed framework be extended from a VLA model to a world action model?
\end{itemize}

Success rate (SR) is used as the primary evaluation metric.
We use LIBERO-PRO \cite{zhou2025libero} as our primary simulation benchmark, as it evaluates generalization to unseen task variations introduced through changes in object layout and appearance, background scene, and task instructions.
For steering experiments, we use the $\pi_{0.5}$ VLA model fine-tuned on the original LIBERO dataset \cite{liu2023libero}, which does not contain demonstrations corresponding to the perturbations introduced in LIBERO-PRO.
\subsection{Effectiveness of Internal Representation Steering}
\label{sec:q1_steering_effectiveness}

\noindent\textbf{Q1: Does steering the internal representations improve task success rates in both simulation and the real world?}

We first evaluate the proposed method using the $\pi_{0.5}$ policy on LIBERO-PRO.
We select 15 LIBERO-PRO tasks for which the unsteered policy achieves a non-saturated success rate between approximately 10\% and 85\%.
This ensures that both successful and failed rollouts are available for identifying outcome-relevant internal representations.
Table~\ref{tab:q1_task_results} reports the success rate of each task with and without steering.
The \emph{Suite} column denotes the original LIBERO task suite together with the LIBERO-PRO perturbation type, where \emph{env} changes the environment or background scene and \emph{swap} changes object positions to create an unseen layout.
Although LIBERO-PRO also includes perturbations such as changes to object appearance and task instructions, we exclude those settings because the $\pi_{0.5}$ policy either fails almost completely or already achieves nearly 100\% success, leaving little room to evaluate steering effectiveness.

\begin{figure}[!t]
      \centering
      \includegraphics[width=0.9\columnwidth]{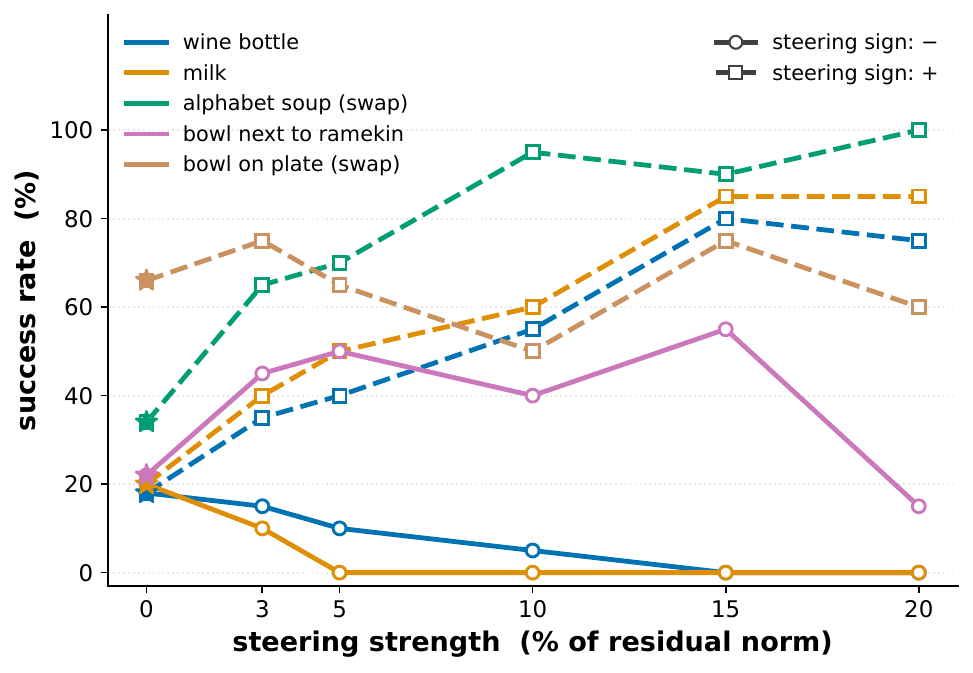}
      \caption{Steering strength vs. SR from a few  tasks to show the typical patterns: monotonic increase/decrease; increase at moderate strengths before degrading at larger strengths.}
      \label{fig:steering_strength}
  \end{figure}

For $\pi_{0.5}$, we apply steering only at the final denoising step and select the action-token position with the highest validation AUC within each transformer layer, as detailed in \Cref{sec:q2_baseline_comparison}.
We sweep each task on $\alpha = [0.03, 0.05, 0.10, 0.15, 0.20]$ for 20 rollouts on one random seed to identify the best steering strength and then use the best steering strength for evaluation for 150 rollouts across 3 random seeds. In \Cref{fig:steering_strength}, we show how the SR changes along with different steering strength from a few representative tasks. 

Overall, steering improves the average success rate from $44.4\%$ to $66.2\%$ on LIBERO-PRO tasks, corresponding to a gain of $21.8$ percentage points.
Among the 15 evaluated tasks, 12 tasks ($80\%$) improve after steering, and all 12 satisfy the criterion that the steered mean SR exceeds the unsteered mean SR plus one standard deviation.
The remaining three tasks do not show a clear improvement in mean SR: L5 and L12 decrease slightly, while L8 remains unchanged.
However, the decreases for L5 and L12 remain within the standard-deviation range of the unsteered performance, suggesting that steering does not cause a substantial degradation on these tasks.

We additionally evaluate RoboIRS on two real-robot manipulation tasks, as illustrated in \Cref{fig:teaser}.
In both tasks, RoboIRS improves the SR of a $\pi_{0.5}$ policy fine-tuned on demonstrations collected under clean-background conditions and evaluated under visually shifted settings.

\definecolor{oursblue}{RGB}{235,243,250}
\definecolor{summarygray}{RGB}{247,247,247}

\begin{table}[t]
    \centering
    \caption{Success rates (\%) of unsteered and steered $\pi_{0.5}$.
    Bold indicates improvements larger than one unsteered standard deviation
    for LIBERO-PRO. The Sign column denotes the steering direction $s$.
    Parentheses in the real-robot rows denote successes/total trials.}
    \label{tab:q1_task_results}

    \scriptsize
    \setlength{\tabcolsep}{2pt}
    \renewcommand{\arraystretch}{1.06}

    \begin{tabularx}{\columnwidth}{
        @{}
        c
        >{\raggedright\arraybackslash}p{1.05cm}
        >{\raggedright\arraybackslash}X
        c
        >{\columncolor{oursblue}\centering\arraybackslash}c
        c
        @{}
    }
        \toprule
        \rowcolor{gray!15}
        \textbf{ID} &
        \textbf{Suite} &
        \textbf{Task} &
        \textbf{Unsteered} &
        \cellcolor{oursblue}\textbf{Steered} &
        \textbf{Sign} \\
        \midrule

        L1 & goal\_env
        & Open top drawer; put bowl inside
        & $12.0 \pm 7.2$
        & $\mathbf{34.0} \pm 2.0$
        & $-1$ \\

        L2 & goal\_env
        & Put bowl on plate
        & $32.7 \pm 1.2$
        & $\mathbf{96.7} \pm 3.1$
        & $-1$ \\

        L3 & goal\_env
        & Put bowl on cabinet
        & $72.0 \pm 7.2$
        & $\mathbf{89.3} \pm 2.3$
        & $-1$ \\

        L4 & goal\_env
        & Put wine bottle on cabinet
        & $17.3 \pm 3.1$
        & $\mathbf{68.0} \pm 2.0$
        & $+1$ \\

        L5 & goal\_swap
        & Put bowl on plate
        & $70.0 \pm 4.0$
        & $66.7 \pm 9.9$
        & $+1$ \\

        L6 & object\_env
        & Alphabet soup to basket
        & $59.3 \pm 1.2$
        & $\mathbf{71.3} \pm 1.2$
        & $-1$ \\

        L7 & object\_env
        & BBQ sauce to basket
        & $80.7 \pm 1.2$
        & $\mathbf{95.3} \pm 2.3$
        & $-1$ \\

        L8 & object\_env
        & Ketchup to basket
        & $84.7 \pm 5.0$
        & $84.7 \pm 2.3$
        & $-1$ \\

        L9 & object\_env
        & Milk to basket
        & $19.3 \pm 1.2$
        & $\mathbf{72.7} \pm 4.6$
        & $+1$ \\

        L10 & obj\_swap
        & Alphabet soup to basket
        & $39.3 \pm 4.6$
        & $\mathbf{90.0} \pm 2.0$
        & $+1$ \\

        L11 & obj\_swap
        & Butter to basket
        & $32.0 \pm 5.3$
        & $\mathbf{43.3} \pm 2.3$
        & $+1$ \\

        L12 & spatial\_env
        & Bowl between plate/ramekin to plate
        & $40.7 \pm 4.2$
        & $36.7 \pm 4.2$
        & $-1$ \\

        L13 & spatial\_env
        & Bowl in top drawer to plate
        & $25.3 \pm 6.1$
        & $\mathbf{34.0} \pm 3.5$
        & $+1$ \\

        L14 & spatial\_env
        & Bowl next to plate to plate
        & $58.7 \pm 4.2$
        & $\mathbf{76.7} \pm 3.1$
        & $-1$ \\

        L15 & spatial\_env
        & Bowl next to ramekin to plate
        & $22.0 \pm 4.0$
        & $\mathbf{33.3} \pm 5.8$
        & $-1$ \\

        \midrule
        \rowcolor{summarygray}
        \multicolumn{3}{r}{\textbf{LIBERO-PRO avg.}}
        & $44.4 \pm 4.0$
        & \cellcolor{oursblue}$\mathbf{66.2} \pm 3.4$
        & -- \\

        \midrule

        H1 & env
        & Disassemble RAM from motherboard
        & 24.0 (12/50)
        & \textbf{44.0 (22/50)}
        & $+1$ \\

        H2 & env
        & Pull out USB connector
        & 12.0 (6/50)
        & \textbf{32.0 (16/50)}
        & $+1$ \\

        \midrule
        \rowcolor{summarygray}
        \multicolumn{3}{r}{\textbf{Real-robot avg.}}
        & 18.0
        & \cellcolor{oursblue}\textbf{38.0}
        & -- \\

        \bottomrule
    \end{tabularx}
\end{table}

\definecolor{RoboBlue}{RGB}{235,244,252}
\definecolor{RoboGray}{RGB}{247,247,247}
\definecolor{RoboHead}{RGB}{239,242,245}
\definecolor{RoboMuted}{RGB}{95,95,95}
\newcolumntype{L}{>{\raggedright\arraybackslash}X}
\newcommand{\rsr}[2]{\ensuremath{#1\,\pm\,#2}}
\newcommand{\rbsr}[2]{\ensuremath{\mathbf{#1}\,\pm\,#2}}
\newcommand{\rtrial}[2]{%
  \begin{tabular}[t]{@{}c@{}}#1\\{\scriptsize(#2)}\end{tabular}}
\renewcommand{\thetable}{\Roman{table}}

\definecolor{bestblue}{RGB}{235,243,250}
\definecolor{baselinegray}{RGB}{247,247,247}

\begin{table}[t]
    \centering
    \caption{Ablation study over steering denoise steps, vectors, and locations,
    ordered by average success rate (SR).}
    \label{tab:steering_strategy}

    \small
    \setlength{\tabcolsep}{4pt}
    \renewcommand{\arraystretch}{1.08}

    \begin{tabularx}{\columnwidth}{@{}Xc@{}}
        \toprule
        \rowcolor{gray!15}
        \textbf{Steering Strategy} & \textbf{Avg. SR (\%)} \\
        \midrule

        One Random Direction             & 44.0 \\

        \rowcolor{baselinegray}
        Unsteered                        & 44.4 \\

        Worst Token per Layer            & 52.7 \\
        Highest-AUC Layer                & 54.3 \\
        Random Token per Layer CAA       & 56.5 \\
        Random Token per Layer           & 57.3 \\
        All Denoise Steps                & 59.3 \\

        \rowcolor{bestblue}
        \textbf{Highest-AUC Token per Layer}
        & \textbf{66.2} \\

        \bottomrule
    \end{tabularx}
\end{table}

\subsection{Comparison with Other Baselines and Ablation Studies}
\label{sec:q2_baseline_comparison}

\noindent\textbf{Q2: Does the proposed approach outperform alternative inference-time intervention methods?}

We compare the proposed approach against other inference-time intervention baselines: contrastive activation addition (CAA)~\cite{rimsky2024steering}, sample and re-rank (Re-rank), and gradient guidance on the flow velocity (Grad.).

This CAA baseline constructs the steering vector using the difference between the residual streams of a selected token from successful and failed rollouts.
For a controlled comparison, the CAA baseline uses the same successful and failed rollout data and intervenes at the same model positions. Steering strengths are selected using the same validation protocol.

Following the same candidate generation and selection paradigm as prior sample and re-rank methods~\cite{nakamoto2025steering,jang2026verifierfree,kwok2025robomonkey}, we implement a re-ranking baseline for our setting.
At each inference step, the observation is replicated across a batch of size $k$, so that a shared forward pass through the vision encoder and multimodal prefix produces $k$ candidate action chunks that differ only in their flow-matching noise.
Each candidate is scored by a learned linear classifier estimating its likelihood of leading to successful execution, and the highest-scoring candidate is executed. We evaluate \(K\in\{8,16\}\) and report \(K=16\), which achieves the higher average SR.

Following the same classifier-guided inference paradigm as prior gradient-guidance methods~\cite{yang2026guided}, we train a two-logit classifier for each task to estimate $P(\mathrm{success}\mid s,a)$ using the mean-pooled PaliGemma layer-17 prefix $s$ and flattened action chunk $a$.
At inference, the policy remains frozen and only the flow velocity is modified.
At each Euler step, we compute a clean-action estimate $\hat{a} = x_t - t\,v_t$, then the flow velocity is modified as
\begin{equation}
v_{\text{guided}}
=
v_t-\alpha P(\text{failure}\mid s,\hat{a})
\frac{\lVert v_t\rVert}{\lVert d\rVert}d,
\end{equation}
where $d \propto \nabla_a \log P(\text{success}\mid s,\hat{a})$. 
The negative sign is because $\pi_{0.5}$ integrates the flow backward in time ($\mathrm{d}t<0$) for the corresponding Euler position update moves in the $\nabla_a \log P(\mathrm{success})$ direction.
We sweep $\alpha \in \{0.02,0.05,0.1,0.2\}$ and use the best value of each task for evaluation.

Table~\ref{tab:baseline_comparison} reports the task-level comparison on the simulation benchmark. Again, each baseline is evaluated on each task for a total of 150 rollouts covering 3 random seeds. 
As shown in Table~\ref{tab:baseline_comparison}, classifier-based steering (RoboIRS) achieves the highest average success rate of $66.2\%$, improving over the unsteered policy by $21.8$ percentage points and outperforming other baselines. This improvement also incurs little additional inference cost in comparison with re-rank which increases the average inference time to 1074.86 ms, approximately 8 times that of other methods. 
Re-ranking outperforms RoboIRS on 7 of 15 tasks by an average margin of 9.7 percentage points, whereas RoboIRS outperforms re-ranking on the remaining 8 tasks by a substantially larger average margin of 22.7 percentage points.
Gradient guidance barely helps in our case because it's found that the classifier relies primarily on the encoded state, producing weak action-space gradients.  
For completeness, computing the guidance gradient with an action-only classifier yields an average SR of 43.5\%.
In contrast, our method remains effective with a simple linear classifier and does not require carefully designed gradient functions.

In addition, we conduct an ablation study to evaluate whether steering the highest-AUC action token in each transformer layer is beneficial for the $\pi_{0.5}$ model.
We compare three intervention strategies: steering all action tokens in the single layer with the highest AUC at the final denoising step, steering the highest-AUC token in each layer across all denoising steps, and steering only the highest-AUC token in each layer at the final denoising step.
As shown in \Cref{tab:steering_strategy}, steering the highest-AUC token in each layer only at the final denoising step achieves the best overall performance for $\pi_{0.5}$.

We also compare classifier-based steering with CAA applied using randomly selected action tokens.
Although CAA can outperform probe-based steering on a few individual tasks, its average SR is lower when the intervention token is not selected using the classifier AUC, indicating that the AUC-based token selection could help. Besides, a random-direction control performs near the unsteered baseline, further supporting the importance of the classifier-derived steering direction.

\begin{table}[t]
\centering
\caption{Comparison of task-level success rates (\%) and average
inference time among baselines and RoboIRS.}
\label{tab:baseline_comparison}
\footnotesize
\setlength{\tabcolsep}{1.5pt}
\renewcommand{\arraystretch}{1.10}
\begin{tabularx}{\columnwidth}{%
  @{}l@{\hspace{2pt}}l
  *{4}{>{\hsize=.97\hsize\linewidth=\hsize\centering\arraybackslash}X}
  >{\hsize=1.12\hsize\linewidth=\hsize\columncolor{RoboBlue}\centering\arraybackslash}X@{}
}
\toprule
\rowcolor{RoboHead}
\multicolumn{2}{@{}l}{\textbf{ID}}
& \shortstack{\textbf{Un-}\\\textbf{steered}}
& \textbf{Grad.}
& \shortstack{\textbf{Re-rank}\\($K=16$)}
& \textbf{CAA}
& \textbf{RoboIRS} \\
\midrule
\multicolumn{2}{@{}l}{L1} & 12.0 & 12.7 & \textbf{44.7} & 34.0 & 34.0 \\
\multicolumn{2}{@{}l}{L2} & 32.7 & 30.7 & 74.0 & 68.7 & \textbf{96.7} \\
\multicolumn{2}{@{}l}{L3} & 72.0 & 58.7 & 79.3 & 75.3 & \textbf{89.3} \\
\multicolumn{2}{@{}l}{L4} & 17.3 & 19.3 & 51.3 & \textbf{94.0} & 68.0 \\
\multicolumn{2}{@{}l}{L5} & 70.0 & 63.3 & 70.7 & \textbf{85.3} & 66.7 \\
\multicolumn{2}{@{}l}{L6} & 59.3 & 56.7 & \textbf{79.3} & 66.0 & 71.3 \\
\multicolumn{2}{@{}l}{L7} & 80.7 & 83.3 & \textbf{98.7} & 95.3 & 95.3 \\
\multicolumn{2}{@{}l}{L8} & 84.7 & 80.0 & \textbf{92.0} & 82.0 & 84.7 \\
\multicolumn{2}{@{}l}{L9} & 19.3 & 20.7 & 32.0 & 16.0 & \textbf{72.7} \\
\multicolumn{2}{@{}l}{L10} & 39.3 & 43.3 & 31.3 & 75.3 & \textbf{90.0} \\
\multicolumn{2}{@{}l}{L11} & 32.0 & 34.7 & 21.3 & \textbf{59.3} & 43.3 \\
\multicolumn{2}{@{}l}{L12} & 40.7 & 43.3 & \textbf{64.0} & 48.0 & 36.7 \\
\multicolumn{2}{@{}l}{L13} & 25.3 & 29.3 & 32.0 & \textbf{36.0} & 34.0 \\
\multicolumn{2}{@{}l}{L14} & 58.7 & 58.0 & 68.0 & 75.3 & \textbf{76.7} \\
\multicolumn{2}{@{}l}{L15} & 22.0 & 20.7 & \textbf{40.7} & 28.7 & 33.3 \\
\midrule
\rowcolor{RoboGray}
\multicolumn{2}{@{}l}{\textbf{Avg. SR}}
& 44.4 & 43.6 & 58.6 & 62.6
& \cellcolor{RoboBlue}\textbf{66.2} \\\midrule
\multirow{2}{*}[-4pt]{\shortstack[l]{Win\\margin}}
& \vline\shortstack{~Base.\\{~}}
& \shortstack{3.7\\{\scriptsize(2/15)}}
& \shortstack{6.6\\{\scriptsize(1/15)}}
& \shortstack{9.7\\{\scriptsize(7/15)}}
& \shortstack{14.8\\{\scriptsize(5/15)}}
& -- \\
& \vline\shortstack{~Ours\\{~}}
& \shortstack{27.8\\{\scriptsize(12/15)}}
& \shortstack{24.6\\{\scriptsize(14/15)}}
& \shortstack{22.7\\{\scriptsize(8/15)}}
& \shortstack{15.9\\{\scriptsize(8/15)}}
& -- \\\midrule
\multicolumn{2}{@{}l}{\shortstack[l]{Inference\\time (ms)}}
& 134.0 & 137.9 & 1074.9 & 136.3 & 136.4 \\
\bottomrule
\end{tabularx}
\par\vspace{3pt}
\parbox{\columnwidth}{\scriptsize
Bold indicates the highest success rate.
Win margins are mean success-rate differences in percentage points on
winning tasks; parentheses report wins/15. Base. denotes the baseline
in each column; Ours denotes RoboIRS. Ties are excluded.}
\end{table}

\subsection{Extension to a World Action Model}
\label{sec:q3_world_action_model}

\noindent\textbf{Q3: Can the proposed framework be extended to a world action model?}

\definecolor{steerblue}{RGB}{235,243,250}

\begin{table}[t]
    \centering
    \caption{Task-level success rates (\%) of the unsteered and steered
    Cosmos Policy. Improvements exceeding one standard deviation of the
    unsteered policy are shown in \textbf{bold}.}
    \label{tab:q3_cosmos}

    \footnotesize
    \setlength{\tabcolsep}{3.5pt}
    \renewcommand{\arraystretch}{1.12}

    \makebox[\columnwidth][c]{%
    \begin{tabularx}{\columnwidth}{
        @{}
        c
        >{\raggedright\arraybackslash}p{1.35cm}
        >{\raggedright\arraybackslash}X
        >{\centering\arraybackslash}p{1.25cm}
        >{\columncolor{steerblue}\centering\arraybackslash}p{1.25cm}
        @{}
    }
        \toprule
        \rowcolor{gray!15}
        \textbf{ID} &
        \textbf{Suite} &
        \textbf{Task} &
        \textbf{Unsteered} &
        \cellcolor{steerblue}\textbf{Steered} \\
        \midrule

        C1 & goal\_object
        & Put bowl on top of cabinet
        & $36.7 \pm 10.0$
        & $\mathbf{92.2 \pm 5.1}$ \\

        C2 & goal\_swap
        & Open top drawer and put bowl inside
        & $16.7 \pm 3.3$
        & $14.4 \pm 5.1$ \\

        C3 & object\_env
        & Pick up alphabet soup and place in basket
        & $17.8 \pm 1.9$
        & $\mathbf{30.0 \pm 3.3}$ \\

        C4 & object\_env
        & Pick up ketchup and place in basket
        & $17.8 \pm 1.9$
        & $\mathbf{24.4 \pm 3.8}$ \\

        C5 & spatial\_swap
        & Pick up black bowl next to plate and place on plate
        & $62.2 \pm 3.8$
        & $\mathbf{97.8 \pm 1.9}$ \\

        C6 & goal\_object
        & Open top drawer and put bowl inside
        & $12.2 \pm 1.9$
        & $\mathbf{27.8 \pm 6.9}$ \\

        C7 & object\_env
        & Pick up BBQ sauce and place in basket
        & $78.9 \pm 6.9$
        & $84.4 \pm 5.1$ \\

        C8 & object\_object
        & Pick up alphabet soup and place in basket
        & $41.1 \pm 5.1$
        & $\mathbf{72.2 \pm 1.9}$ \\

        \midrule
        \rowcolor{gray!8}
        \multicolumn{3}{r}{\textbf{Average}}
        & 35.4
        & \cellcolor{steerblue}\textbf{55.4} \\

        \bottomrule
    \end{tabularx}%
    }
\end{table}

To determine whether the method is specific to the internal representations of $\pi_{0.5}$ policy, we additionally apply it to Cosmos Policy \cite{kim2026cosmos} which has a different model architecture (DiT) and a different training recipe.
Eight tasks with unsaturated SR in LIBERO-PRO is selected for evaluation.
We sweep the each task on $\alpha = [0.05, 0.10, 0.15, 0.30]$ for 10 rollouts on one random seed to identify the best steering strength per task. Then evaluate on a total of 90 rollouts over 3 random seeds.
As shown in Table~\ref{tab:q3_cosmos}, steering improves the macro-average success rate of Cosmos Policy from $35.4\%$ to $55.4\%$.
Among the eight evaluated tasks, six achieve meaningfully higher success rates after steering.

However, Cosmos Policy differs substantially from $\pi_{0.5}$ in its action-generation mechanism. It does not use a separate action expert.
Instead, it repeatedly tiles an action chunk into a latent frame and generates that frame with the video diffusion transformer. The action is decoded by averaging these repeated representations. 
In the LIBERO-PRO configuration, the action frame contains 196 tokens. 
Steering a single token therefore affects only a small part of the action latent. We instead steer all 196 action-frame tokens in the final transformer block. 
The corresponding classifier is instead trained on mean-pooled residual streams over the 196 tokens.
Because the classifier is trained in normalized residual-stream space, there are two choices of steering vector $\mathbf{w}/\boldsymbol{\sigma}$ and $\mathbf{w}\odot\boldsymbol{\sigma}$.
Interpreting $\mathbf w$ as the classifier normal gives $\mathbf w/\boldsymbol{\sigma}$, which corresponds to the gradient of the classifier score with respect to the raw residual.
Alternatively, interpreting $\mathbf w$ as a displacement direction in normalized representation space gives $\mathbf w\odot\boldsymbol{\sigma}$ after mapping the displacement back to the raw residual coordinates.
Empirically, for Cosmos Policy, $\mathbf{w}/\boldsymbol{\sigma}$ performs better on tasks C1--C5, whereas $\mathbf{w}\odot\boldsymbol{\sigma}$ performs better on C6--C8.
Overall, these results show that RoboIRS transfers beyond $\pi_{0.5}$ to a policy with a different architecture and training recipe, while requiring architecture-specific choices for classifier and intervention sites.

\subsection{Observations and Analysis}
\label{sec:q4_why_when_steering}

In this subsection, we share a few observations and analysis by answering the following questions.

\begin{figure}[!t]
    \centering

    \begin{subfigure}{\columnwidth}
        \centering
        \includegraphics[width=\linewidth]{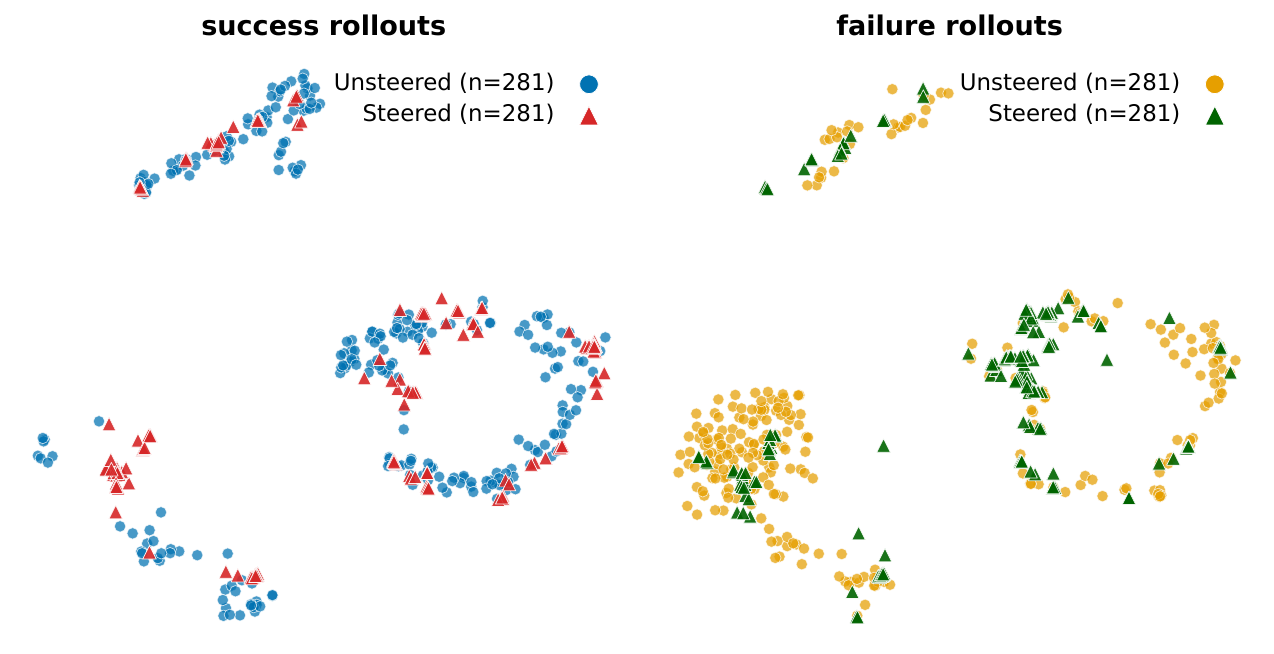}
        \caption{alphabet soup (L10) task.}
    \end{subfigure}

    \begin{subfigure}{\columnwidth}
        \centering
        \includegraphics[width=0.9\linewidth]{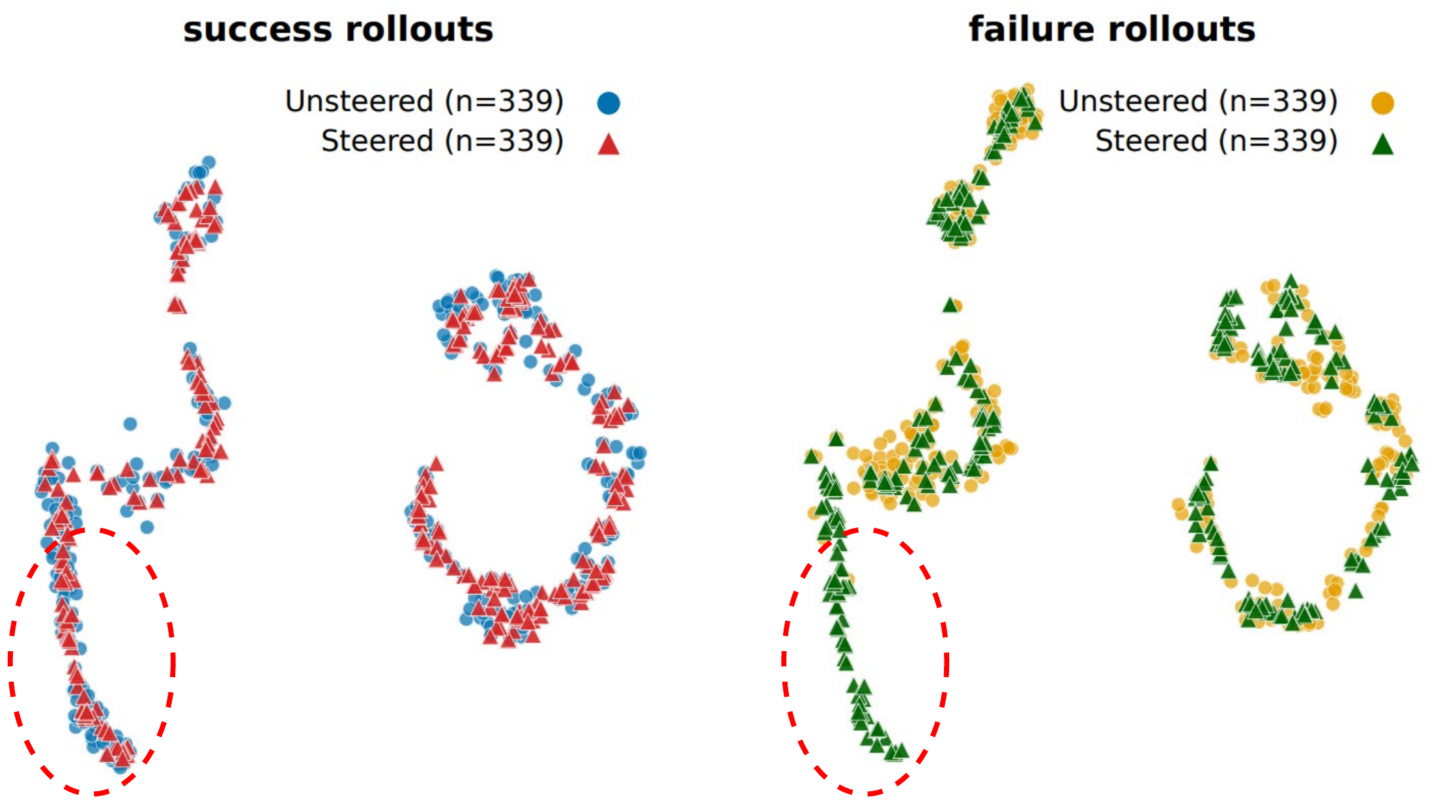}
        \caption{ketchup (L8) task.}
    \end{subfigure}

    \caption{t-SNE visualization of the last-layer's residuals. Each panel compares residual representations from unsteered and steered policies separately for successful and failed rollouts.}
    \label{fig:t-sne}
\end{figure}

\textbf{How do residual-stream representations change before and after steering?}
Steering tends to make the residual representations more locally concentrated for most tasks, although this effect is not consistent across all tasks.
We visualize the residual streams from unsteered and steered rollouts using t-SNE~\cite{t-sne}.
As shown in \Cref{fig:t-sne}(a), for the strongly improved L10 task, the steered residuals form tighter local clusters while remaining globally distributed over a similar region as the unsteered residuals.
In \Cref{fig:t-sne}(b), however, some steered residuals from failed L8 rollouts appear in regions occupied by successful rollouts but not by unsteered failed rollouts, even though steering does not improve the task-level SR.
This might suggest that a direction predictive of success and failure may recover some success-like residual representations without necessarily providing a causally effective steering direction for improving task SR.

\textbf{What types of failure modes are corrected by steering?}
We observe two broad categories of failure modes, consistent with the failure taxonomy discussed in Sentinel~\cite{AgiaSinhaEtAl2024}.
The first is dominated by decision-making errors, including L3, L6, L9, L10, L11, L14, and L15, where the policy often makes an incorrect high-level choice despite relatively accurate low-level execution.
For example, in L10, the unsteered policy frequently selects butter instead of alphabet soup after the object-position swap, whereas the steered policy usually selects the correct object.
Across these tasks, RoboIRS improves the average SR by $24.9\%$.

The second category is dominated by action-precision errors, including L1, L2, L4, L5, L7, L8, L12, and L13.
Typical failures include inaccurate grasps that cause objects to slip, repeated unsuccessful execution leading to timeout, or the gripper becoming stuck on the drawer handle during opening.
For these tasks, the average SR improves by $19.1\%$.

For the three tasks where steering does not improve performance, the corresponding classifiers also have lower validation AUCs: 0.701 for L5, 0.827 for L8, and 0.820 for L12, compared with an average AUC of 0.90 across the selected steering locations for all tasks.

\textbf{Can a steering direction transfer across perturbations of the same task?}
Not necessarily, transfer appears to depend on whether the perturbations induce similar failure reasons.
For ``Put the bowl on the plate,'' transferring the L2 steering vector to L5 improves SR from $62\%$ to $86\%$ on the same evaluation seed. Both tasks commonly fail by grasping the bowl inaccurately.
In contrast, transferring the L10 vector to L6 for ``Pick up the alphabet soup and place it in the basket'' reduces SR to $46\%$.
In L10, the robot mainly confuses alphabet soup with butter, whereas it confuses it with salad dressing in L6, suggesting that the learned directions may encode different failure-specific concepts.

%% file: tex/conclusion.tex
\section{Conclusion and Limitations}

\textbf{Conclusion.}
We propose RoboIRS, an inference-time representation-steering method that uses successful and failed rollouts to identify and manipulate outcome-relevant internal representations without updating policy parameters.
RoboIRS improves $\pi_{0.5}$ performance on LIBERO-PRO from $44.4\%$ to $66.2\%$, also improves real-robot manipulation, and transfers to Cosmos Policy, where success rate increases from $35.4\%$ to $55.4\%$.
These results show that internal representations of frozen robot policies can be directly steered to improve task performance at inference time.

\textbf{Limitations.}
The proposed method may be less effective on tasks where successful and failed rollouts do not exhibit clear differences in their actions, making it difficult to label and train a classifier.
Our experiments are also limited to a small number of VLA and WAM policy architectures, as the primary goal of this work is to investigate the feasibility of transferring representation-steering techniques from LLMs to robotic policies.
Different model architectures may encode task-relevant information differently and may therefore require different intervention locations, steering vectors, or steering strategies.
Finally, like related representation-steering methods for LLMs, RoboIRS requires empirical selection of hyperparameters such as steering strength, introducing additional cost before deployment.